\documentclass[11pt]{article}
\usepackage{coling2018}
\usepackage{times}
\usepackage{amsmath}
\usepackage{bbm}
\usepackage{latexsym}
\usepackage{pbox}
\usepackage{graphicx}
\usepackage{longtable}
\usepackage{answers}
\usepackage{array}
\usepackage{hyperref}
\usepackage{xcolor}
\usepackage{booktabs}
\usepackage{ctable}
\definecolor{advcolor}{RGB}{2, 0, 136}
\newcommand*{\affmark}[1][*]{\textsuperscript{#1}}
\newcommand*{\affaddr}[1]{#1}

\title{Stress Test Evaluation for Natural Language Inference}

\author{Aakanksha Naik\affmark[1]\thanks{*The first two authors contributed equally to this work.} , Abhilasha Ravichander\affmark[1]\footnotemark[1] , \\ \textbf{Norman Sadeh}\affmark[2],   \textbf{Carolyn Rose}\affmark[1], \textbf{Graham Neubig\affmark[1]} \\
\affaddr{\affmark[1]Language Technologies Institute, Carnegie Mellon University}\\
\affaddr{\affmark[2]Institute of Software Research, Carnegie Mellon University}\\
  {\tt\{anaik, aravicha, sadeh, cprose, gneubig\}@cs.cmu.edu}}

\begin{document}
\maketitle
\begin{abstract}
Natural language inference (NLI) is the task of determining if a natural language hypothesis can be inferred from a given premise in a justifiable manner. NLI was proposed as a benchmark task for natural language understanding. Existing models perform well at standard datasets for NLI, achieving impressive results across different genres of text. However, the extent to which these models understand the semantic content of sentences is unclear. In this work, we propose an evaluation methodology consisting of automatically constructed ``stress tests'' that allow us to examine whether systems have the ability to make real inferential decisions. Our evaluation of six sentence-encoder models on these stress tests reveals strengths and weaknesses of these models with respect to challenging linguistic phenomena, and suggests important directions for future work in this area.  
\end{abstract}

\section{Introduction}

\blfootnote{
    %
    % for review submission
    %
    % final paper: en-us version 
    
    \hspace{-0.65cm}  % space normally used by the marker
    This work is licensed under a Creative Commons 
    Attribution 4.0 International License.
    License details:
    \url{http://creativecommons.org/licenses/by/4.0/}
}

%Sentence encoding models motivate
%motivate multinli

%\gn{The intro is now a bit too long. Try to simplify the wording and remove unnecessary info.}

%Motivate speicfic to entailment

Natural language inference (NLI), also known as recognizing textual entailment (RTE), is concerned with determining the relationship between a premise sentence and an associated hypothesis. This requires a model to make the 3-way decision of whether a hypothesis is true given the premise (\emph{entailment}), false given the premise (\emph{contradiction}), or whether the truth value cannot be determined (\emph{neutral}). NLI has been proposed as a benchmark task for natural language understanding research \cite{cooper1996using,dagan2006pascal,giampiccolo2007third,dagan2013recognizing,snli,nangia-EtAl:2017:RepEval}, due to the requirement for models to reason about several difficult linguistic phenomena\footnote{such as scope, coreference, quantification, lexical ambiguity, modality and belief.} to perform well at this task \cite{snli,williams2017broad}.
%\gn{Make sure you always have a space before citations.}. 
%Sentence-encoder models for inference, in particular, have been considered to be of great import to natural language understanding due to their ability to represent sentences as fixed length vectors and perform reasoning based on these vector representations\cite{nangia2017repeval}.

% \gn{maybe explain about how these datasets were created very very briefly
Due to its importance, significant prior work \cite{dagan2006pascal,dagan2009recognizing,dagan2013recognizing,marelli-EtAl:2014:SemEval} has focused on developing datasets and models for this benchmark task. Most recently, this task has been concretely implemented in the Stanford NLI (SNLI; \newcite{snli}), and Multi-genre NLI (MultiNLI; \newcite{williams2017broad}) datasets, where crowdworkers are given a premise sentence and asked to generate novel sentences representing the three categories of entailment relations. Within the scope of these benchmark datasets, state-of-the-art deep learning-based sentence encoder models\footnote{Sentence-encoder models are considered to be especially important to natural language understanding due to the requirement to represent sentences as fixed length vectors and perform reasoning based on these representations \cite{nangia-EtAl:2017:RepEval}. Thus, they will be our primary focus.} (\newcite{nie-bansal:2017:RepEval}, \newcite{chen-EtAl:2017:RepEval}, \newcite{conneau-EtAl:2017:EMNLP2017}, \newcite{balazs-EtAl:2017:RepEval}, \emph{inter alia}) have been shown to consistently achieve high accuracies, which may lead us to believe that these models excel at  NLI across genres of text. However, machine learning models are known to exploit idiosyncrasies of how the data was produced, allowing them to imitate desired behavior using tricks such as pattern matching \cite{levesque2014our,rimell2009unbounded,papernot2017practical}. In NLI this arises when the test set contains many easy examples, and the extent to which difficult components of language understanding are required is masked within traditional evaluation. Therefore we ask a natural question: is good model performance on NLI benchmarks a result of sophisticated pattern matching, or does it reflect true competence at natural language understanding?
In this work, we propose an evaluation based on ``stress'' tests for NLI, which tests robustness of NLI models to specific linguistic phenomena. This methodology is inspired by the work of \newcite{jia-liang:2017:EMNLP2017} (and other related work: \S\ref{ref:Discussion}), which proposes the use of adversarial evaluation for reading comprehension by adding a distracting sentence at the end of a paragraph (known as concatenative adversaries), and evaluating models on this test set. However, this evaluation scheme cannot easily be applied to NLI as i) the adversarial perturbations suggested are not \emph{label preserving}, that is the semantic relation between premise and hypothesis will not be maintained by such edits, ii) the perturbations may decrease system performance but are not interpretable and iii) premise-hypothesis pairs in inference usually consist of a single sentence, but concatenative adversaries break this assumption. In addition, besides evaluating model robustness to \emph{distractions} in the form of adversarial examples, we are also interested in evaluating model \emph{competence} on types of reasoning necessary to perform well at the task.
% In this work, we propose an evaluation based on ``stress'' tests for NLI, which tests robustness of NLI models to specific linguistic phenomena. This methodology is inspired by the work of \newcite{jia-liang:2017:EMNLP2017} (and other related work: \S\ref{ref:Discussion}), which proposes the use of adversarial evaluation for reading comprehension by adding a distracting sentence at the end of a paragraph (known as \emph{concatenative adversaries}), and evaluating models on this test set. However, this evaluation scheme cannot easily be applied to NLI as i) The adversarial perturbations suggested are not label preserving, that is the semantic relation between premise and hypothesis will not be maintained by such edits, ii) the perturbations may decrease system performance but are not interpretable and iii) premise-hypothesis pairs in inference usually consist of a single sentence, but concatenative adversaries break this assumption. 
 %\gn{Is this the only reason why? This seems like a quite trivial problem to overcome. I'm sure that there are other reasons!}.
% It is unclear how to construct semantic-preserving adversarial examples for inference.
% \gn{``fine-grained'' is no problem, as we indeed have fine-grained error categories.}
% , because progress on this task is considered to be essential to natural language understanding.

The proposed method offers a fine-grained evaluation for NLI (with label-preserving adversarial perturbations when required), in the form of ``stress tests''. In the stress testing methodology, systems are tested beyond normal operational capacity in order to identify weaknesses and to confirm that intended specifications are being met \cite{Hartman:1967:WSS:1465611.1465713,tretmans1999testing,beizer2003software,pressman2005software,nelson2009accelerated}.  To construct these stress tests for NLI, we first examine the predictions of the best-performing sentence encoder model on MultiNLI \cite{nie-bansal:2017:RepEval}, and create a typology of phenomena that it finds challenging (\S\ref{section:NLI}). Based on this typology, we present a methodology to automatically construct stress tests, which may cause models that suffer from similar weaknesses to fail (\S\ref{section:Construction}).  The resulting tests make it possible to perform evaluation on a phenomenon-by-phenomenon basis, which is not the case for \newcite{jia-liang:2017:EMNLP2017}%
\footnote{\newcite{isabelle-cherry-foster:2017:EMNLP2017} have proposed a similar fine-grained evaluation approach for machine translation, but it requires the manual construction of examples, unlike our automatic approach.}. We benchmark the  performance of four state-of-the-art models on the MultiNLI dataset on our constructed stress tests (\S\ref{section:experiments}), and observe performance drops across stress tests. We view these results as a first step towards robust, fine-grained evaluation of NLI systems. To encourage development of models that perform true natural language understanding for NLI, we release our code and all
stress tests for future evaluation\footnote{All stress tests and resources available at \url{https://abhilasharavichander.github.io/NLI_StressTest/}}.

\section{Weaknesses of State-of-the-art NLI}
\label{section:NLI}

\begin{table*}[t!]
\small
\centering
\begin{tabular}{|p{2cm}|p{6.5cm}|p{6.5cm}|}
\hline
\textbf{Error} & \textbf{Premise} & \textbf{Hypothesis} \\ \hline
 \textbf{Word Overlap} (N$\rightarrow$E) & And, could it not result in a decline in Postal Service volumes across--the--board? & There may not be a decline in Postal Service volumes across--the--board.\\ \hline
\textbf{Negation} (E$\rightarrow$C) & Enthusiasm for Disney's Broadway production of The Lion King dwindles. & The broadway production of The Lion King is no longer enthusiastically attended. \\ \hline
\textbf{Numerical Reasoning} (C$\rightarrow$E)& Deborah Pryce said Ohio Legal Services in Columbus will receive a \$200,000 federal grant toward an online legal self-help center. & A \$900,000 federal grant will be received by Missouri Legal Services, said Deborah Pryce. \\ \hline
\textbf{Antonymy} (C$\rightarrow$E) & ``Have her show it," said Thorn. & Thorn told her to hide it.\\ \hline
\textbf{Length} \textbf{Mismatch} (C$\rightarrow$N) & So you know well a lot of the stuff you hear coming from South Africa now and from West Africa that's considered world music because it's not particularly using certain types of folk styles. & They rely too heavily on the types of folk styles.\\ \hline
\textbf{Grammaticality} (N$\rightarrow$E) & So if there are something interesting or something worried, please give me a call at any time. & The person is open to take a call anytime. \\ \hline
\textbf{Real World Knowledge} (E$\rightarrow$N) & It was still night. & The sun hadn't risen yet, for the moon was shining daringly in the sky.\\ \hline
\textbf{Ambiguity} (E$\rightarrow$N) & Outside the cathedral you will find a statue of John Knox with Bible  in hand. & John Knox was someone who read the Bible. \\ \hline
\textbf{Unknown} (E$\rightarrow$C) & We're going to try something different this morning, said Jon.& Jon decided to try a new approach. \\ \hline
\end{tabular}
\label{tab:errorexfinal}
\caption{Examples of misclassified samples for each error category (Gold Label$\rightarrow$Predicted Label) }
\end{table*}

% Since NLI is traditionally considered a first step towards deeper and more general understanding of language, it is important to study the challenges faced by current state-of-the-art inference models on the NLI task.  
%Wasn't sure if this tied in, we might need to reframe this part: For NLI datasets, if the hypothesis seems very similar to the premise, it should be a clear case for entailment, however if it introduces even a little extra information, it should actually belong to the neutral class. However, if the extra information introduced in the hypothesis can be very easily inferred from the premise, it might still be considered entailment. This distinction requires the model to reason about what a human can reasonably infer, and is actually much harder to differentiate between. This characterization can help models to selectively penalize by error type during training.
%\gn{Is there any previous work that we can cite here? It seems like there should be some on RTE before neural models. Could you survey a little and add the results to the appropriate bullets below?}
Before creating adversarial examples that challenge state-of-the-art systems, it is helpful to study what phenomena systems find difficult. To elucidate this, we conduct a comprehensive error analysis of the best-performing sentence encoder model for MutliNLI \cite{nie-bansal:2017:RepEval}.
%\begin{figure}
%\centering
% \includegraphics[scale=0.18]{final_err_analysis.pdf}
% \caption{Various possible sources of error in the matched development set}
% \label{fig:results1}
%\end{figure}
% \gn{This is probably not necessary}
% The matched development set contains examples from genres of text which are also present in the MultiNLI training set, while the mismatched development set contains examples from previously unseen genres of text. 
%\ar{Confusion matrix for baselines} 
%The confusion matrices for both sets are shown in Tables \ref{tab:errormis} and \ref{tab:errormatch}. It is interesting to observe that for both, the matched and mismatched development sets, the largest category of errors comes from overpredicting entailment.
For this analysis, we sample 100 misclassified examples from both genre-matched and mismatched sets, analyze their potential sources of errors, and group them into a typology of common reasons for error.%
% \footnote{Error analysis samples and annotations are provided in the supplementary material.}
%\gn{I would prefer to keep the pie chart. It's nice to have a colorful and attractive figure in the paper to spice it up a little and prevent people from being overwhelmed by a wall of text. I can give suggestions for other places to reduce the length.}
 In the end, the reasons for errors can broadly be divided into the following categories,\footnote{\% indicates what proportion of examples from our error analysis on the matched development set, fall into each category. Results on the mismatched development set follow similar trends and are available in appendix A} with examples shown in Table 1:
%\gn{I think this should be ``the reasons for errors''. This probably needs to be fixed throughout.} broadly consist of the following types, with examples shown in Table \ref{tab:errorex}:
\begin{enumerate}
\item \textbf{Word Overlap (29\%)}: Large word-overlap between premise and hypothesis sentences causes wrong entailment prediction, even if they are unrelated. Very little word overlap causes a prediction of neutral instead of entailment. 
\item \textbf{Negation (13\%)}: Strong negation words (``no'', ``not'') cause the model to predict contradiction for neutral or entailed statements. 
%This results in Type I and Type II errors.  
%This results in Type III and Type IV errors.
\item \textbf{Antonymy (5\%)}: Premise-hypothesis pairs containing antonyms (instead of explicit negation) are not detected as contradiction by the model.
\item \textbf{Numerical Reasoning (4\%)}: For some premise-hypothesis pairs, the model is unable to perform reasoning involving numbers or quantifiers for correct relation prediction.
\item \textbf{Length Mismatch (3\%)}: The premise is much longer than the hypothesis and this extra information could act as a distraction for the model.
\item \textbf{Grammaticality (3\%)}: The premise or the hypothesis is ill-formed because of spelling errors or incorrect subject-verb agreement.
% \item \textbf{Grammaticality (3\%)}: In these instances, the premise or the hypothesis (or both) contain misspellings due to either Turker error or noise in the source text.
\item \textbf{Real-World Knowledge (12\%)}: These examples are hard to classify without some real-world knowledge.
\item \textbf{Ambiguity (6\%)}: For some instances, the correct answer is unclear to humans. These are the most difficult cases.
\item \textbf{Unknown (26\%)}: No obvious source of error is discernible in these samples. 
\end{enumerate}
Some of our error categories such as real world knowledge are well-known ``hard'' problems. However, error categories such as negation scope and antonymy, are crucial for natural language understanding and been of significant interest to study in formal semantics \cite{kroch1974semantics,muehleisen1997antonymy,murphy2003semantic,moscati2006scope,brandtler2006aristotle}. Some of these phenomena have long been suspected to be challenging for entailment models \cite{jijkoun2006recognizing,lobue2011types,roy2017reasoning}. In fact, at roughly the same time as this submission, \newcite{gururangan2018annotation} corroborate our findings by identifying lexical choice such as negation words, as well as sentence length as biasing factors in NLI datasets. 

\section{Stress Test Set Construction}
% \an{Fleiss Kappa Agreements on Label: Length Mismatch (0.78), Word Overlap (0.79), Quant (0.75), Antonym (), Negation ()}\\\\
\label{section:Construction}
\begin{table*}[h]
\centering
\resizebox{\textwidth}{!}{%
\begin{tabular}{|p{2.0cm}|p{7.5cm}|p{7.5cm}|}
\hline \textbf{Error Cat.} & \textbf{Premise} & \textbf{Hypothesis} \\ \hline
\textbf{Antonyms} & I \textcolor{advcolor}{love} the Cinderella story. & I \textcolor{advcolor}{hate} the Cinderella story.\\ \hline
\textbf{Numerical Reasoning} & \textcolor{advcolor}{Tim has 350 pounds of cement in 100, 50, and 25 pound bags} & \textcolor{advcolor}{Tim has less than 750 pounds of cement in 100, 50, and 25 pound bags} \\ \hline
\textbf{Word \newline Overlap} & Possibly no other country has had such a turbulent history. & The country's history has been turbulent \textcolor{advcolor}{and true is true}  \\ \hline
\textbf{Negation} & Possibly no other country has had such a turbulent history. & The country's history has been turbulent \textcolor{advcolor}{and false is not true}\\ \hline
\textbf{Length \newline Mismatch} & Possibly no other country has had such a turbulent history \textcolor{advcolor}{and true is true and true is true and true is true and true is true and true is true} & The country's history has been turbulent. \\ \hline
% \textbf{Antonyms} & The flame was easy to control, but lacked coldness. & The flame was easy to control, but lacked heat. & C\\ \hline
\textbf{Spelling \newline Errors}  & As he emerged, Boris remarked, glancing  up at \textcolor{advcolor}{teh} clock: "You are early & Boris had just arrived at the rendezvous when he appeared  \\ \hline
\end{tabular}
}
\caption{Example constructions from stress tests}
\label{table:stresstestexample}
\end{table*}

While the error analysis is informative for \newcite{nie-bansal:2017:RepEval}, performing a manual analysis for every system is not scalable. To create an automatically calculable proxy, we focus on automatically constructing large-scale datasets (\emph{stress tests}), which test NLI models on phenomena that account for most errors in our analysis. In particular, we generate adversarial examples which test ``word overlap'', ``negation'', ``length mismatch'', ``antonyms'', ``spelling error'' and ``numerical reasoning''.%
\footnote{Notably, we focus only on ``spelling error'' for ``grammaticality''. We omit ``real world knowledge'' as it is not trivial to create a large dataset without human input, the ``ambiguity'' category  because it is unreasonable that models can handle such cases, and the ``unknown'' category because it does not correspond to a particular phenomenon.}
%This indicates that the current evaluation framework for inference does not rigorously test the capacity of models to capture complex phenomena \gn{I don't think this is true. The current model is paying a price for not getting these right. I'd be very careful about wording here.}.
%\gn{Why these six? Also, don't capitalize things like the ones in this sentence such as ``Word overlap''. You should only capitalize proper names.}.

% \gn{Made some modifications here (original commented below. Main ones are ``reasoning'' becomes ``lexical reasoning'', as usually reasoning indicates things much more complicated than numbers and antonyms.}It contains tests for ``antonymy'' and ``numerical reasoning''.
We organize our stress tests into three classes, based on their perceived difficulty for the model. The first class (\emph{competence tests}) evaluates the model's ability to reason about quantities and understand antonymy relations.  The second class (\emph{distraction tests}), estimates model robustness to shallow distractions such as lexical similarity or presence of negation words. This category contains ``word overlap'', ``negation'' and ``length mismatch'' tests. The final class (\emph{noise tests}) checks model robustness to noisy data and consists of our ``spelling error'' test. Our adversarial construction uses three techniques: heuristic rules with external knowledge sources (for competence tests), a propositional logic framework (for distraction tests) and randomized perturbation (for noise tests). The following subsections describe our stress test construction, with examples shown in Table \ref{table:stresstestexample}.

\subsection{Competence Test Construction}
% The first class of stress tests examine the model's ability to reason about ``antonymy'' and ``numerical reasoning'', and are constructed using heuristic rules in conjunction with lexical resources.\\
\noindent
\textbf{Antonymy:} For this construction, we consider every sentence from premise-hypothesis pairs in the development set independently. We perform word-sense disambiguation for each adjective and noun in the sentence using the Lesk algorithm \cite{Lesk:1986:ASD:318723.318728}. We then sample an antonym for the word from WordNet \cite{miller1995wordnet}. The sentence with the word substituted by its antonym and the original sentence become a new premise-hypothesis pair in our set. This results in 1561 and 1734 premise-hypothesis pairs for matched and mismatched sets respectively.

Substituting a word with its antonym may not always result in a contradiction \footnote{Consider examples of sentences with modalities, belief, conjunction or even conversational text such as \emph{``They can change the tone of people's voice yes.''},
\emph{``They can change the tone of people's voice no.''}, coreference, word substitution in metaphors or failure of word sense disambiguation.} . 
Hence, three annotators were provided 100 random samples from the stress test set to evaluate for correctness. At least two annotators agreed on 86\% of the labels being contradiction. We also evaluated grammaticality of our constructions, with at least two annotators agreeing on 87\% being grammatical.
%\\ % \gn{100% agreement of two of three annotators on a binary decision is a tautology...} of the sentences and at least two agreeing on 100\%.\\

\noindent
\textbf{Numerical Reasoning:}
Creating a stress test for numerical reasoning is non-trivial as most of the MultiNLI development set does not involve quantities. Hence, we extract premise sentences from AQuA-RAT \cite{ling-EtAl:2017:Long}, a dataset specifically focused on algebraic word problems along with rationales for their solutions. However, word problems from AQuA-RAT are quite complicated\footnote{Including concepts such as probability, geometry and theoretical proofs.} and general-purpose NLI models cannot reasonably be expected to solve them.
%Hence, we perform some preprocessing to filter out such ``hard'' samples and generate a reasonable set of premise sentences. 

To generate a reasonable set of premise sentences, we first discard problems which do not have numerical answers or have long rationales ($>$3 sentences) as such problems are inherently complex. We then split all problems into individual sentences and discard sentences without numbers, resulting in a set of 40,000 sentences. From this set, we discard sentences which do not contain at least one named entity (we consider ``PERSON'', ``LOCATION'' and ``ORGANIZATION''), since such sentences mostly deal with abstract concepts.%
\footnote{For example, ``Find the smallest number of five digits exactly divisible by 22, 33, 66 and 44''.}
This results in a set of 2500 premise sentences. For each premise, we generate entailed, contradictory and neutral hypotheses using heuristic rules:\\
1. \textbf{Entailment:} Randomly choose and change one numerical quantity from the premise, prefixing it with the phrase ``less than'' or ``more than'' based on whether the new number is higher or lower.\\
2. \textbf{Contradiction:} Perform one of two actions with equal probability: randomly choose a numerical quantity from the premise and change it, or randomly choose a numerical quantity from the premise and prefix it with ``less than/ more than'' without changing it.\\
3. \textbf{Neutral:} Flip the corresponding entailed premise-hypothesis pair. \\
%\par
Using these rules, we generate a set of 7,596 premise-hypothesis pairs testing models on their ability to perform numerical reasoning. We further instruct three human annotators to evaluate 100 randomly sampled examples for difficulty, grammaticality and label correctness (since the labels are automatically generated). At least two annotators agreed with our generated label for 91\% of the samples. Additionally, at least two annotators agreed on 92\% of the examples being grammatical, and 98\% being trivial numerical reasoning for humans. % \footnote{Atleast two annotators agreed on difficulty in 98\% of the cases}.
% \an{ Quant: annotators agree on 98, match with gold on 91}\\
% \an{, Quant: annotators agree on 100, say grammatical to 92, for ease of quant: annotators agree on 98, say easy to 98}

\subsection{Distraction Test Construction}
This class includes stress tests for ``word overlap'', ``negation'' and ``length mismatch'', which test model ability to avoid getting distracted by simple cues such as lexical similarity or strong negation words. Models usually exploit such cues to achieve high performance since they have strong but spurious correlations with gold labels, but reliance on shallow reasoning can be used to distract models easily, as we demonstrate. We use a framework inspired by propositional logic to construct adversarial examples.\\

\noindent 
\textbf{Propositional Logic Framework:} Assume a premise $p$ and a hypothesis $h$. For entailment, $(p \Rightarrow h) $ $\implies (p \land True \Rightarrow h)$ since  ($ p \land True = p$). Similarly for contradiction, 
  $(p \Rightarrow\!\Leftarrow h) $ $\implies (p \land True \Rightarrow\!\Leftarrow h)$ and if $p$ and $h$ are neutral, they remain neutral. In other words, if the premise or hypothesis is in conjunction with a statement that is independently true in all worlds, the entailment relation is preserved.  

The next step is to construct such statements whose values are true in all worlds (\emph{tautologies}). We then define stress test accuracy for NLI as :
$$Adv(a) = \frac{1}{|D_{test}|}\sum\limits_{i \in D_{test}}\mathbbm{1}\{f(p,h, a)=c_{i}\}$$ where $a$ is an adversarial tautology which can be attached as a conjunction to either the premise or hypothesis without changing the relation. We use this framework to construct distraction tests for ``word overlap'', ``negation'' and ``length mismatch''. For all sets, we use simple tautologies, which do not contain words that share any topical significance with the premise or hypothesis.

A natural concern is that statements obtained from such constructions are unnatural \cite{grice1975logic}, making the NLI task more difficult for humans. To study this, we run a human evaluation where three annotators are shown premise-hypothesis pairs from these sets and instructed to label the relation. On word overlap, we find the provided label has 91\% agreement with the gold label. For length mismatch, the provided label has 85\% agreement with gold. This is similar to the agreement reported in \newcite{williams2017broad}, leading us to believe the constructed examples are not too unnatural or difficult. The constructions also remain grammatical; after annotating 100 samples from our adversarially generated set, only two were deemed ungrammatical, and both were because of reasons unrelated to our perturbations.
%\footnote{(' no one from the white hosue was interrogated regarding the murders and false is not true.' and ` nah, everyone reporter i've known has always treated victims with respect and false is not true.')} (the sentences without the adversarial additions are in themselves ungrammatical) .
Specific details for our sets are as follows:
%This aligns with our expectations as a tautology can be used in conjunction with a sentence, and as long as the tautology and the sentence are both valid, the conjunction will be valid as well.
%\gn{I think the biggest concern about this section is that these result in very very unnatural sentences (even if they are grammatical). I think you should do something somewhere in this section to assuage the reader's concerns that these sentences are not relevant.}. 
%(which entailment models cannot be expected to understand)

\noindent
\textbf{Word Overlap:} For this set, we append the tautology \emph{``and true is true''} to the end of the hypothesis sentence for every example in the MultiNLI development set.\\ 
%In this test set, we construct our adversarial examples in stages. We gradually add longer tautologies to the hypothesis, to demonstrate the effects of decreasing word overlap between the premise and they hypothesis. Beginning with a simple tautology \emph{"and true"} (\textsc{taut1}), which is true in all worlds we graduate to longer tautologies such as "and true is true"(\textsc{taut2}), "and true is true and true is true"(\textsc{taut3}). These statements are still true in all possible worlds and do not contain words that share any topical significance with the premise and the hypothesis. Additionally, we add them to the end of the sentence to avoid topical bias.
%We run an additional sanity check study where we annotate 100 adversarial examples with the relation between them. This is to measure if the adversarial samples are difficult for humans to categorize as well, indicated that either the semantics has changed or the sentence has become too complex making determining the relation difficult. We categorize 97 of the 100 relations correctly as per the gold label, which is much higher than the inter-annotator agreement achieved for the MultiNLI dataset\cite{williams2017broad}. 
\noindent
\textbf{ Negation:} For this set, we append the tautology \emph{``and false is not true''}, which contains a strong negation word (\emph{``not''}), to the end of the hypothesis sentence for every example in the MultiNLI development set.\\  %ADD ANNOTATION RESULTS 
\noindent 
\textbf{Length Mismatch:} For this adversarial set, we append the tautology \emph{``and true is true''} five times to the end of the premise sentence for every example in the MultiNLI development set. We modify the premise sentence in this case as we hypothesize that errors in this category mainly arise due to the premise sentence being unwieldy.  %ADD ANNOTATION RESULTS

\subsection{Noise Test Construction}
This class consists of an adversarial example set which tests model robustness to spelling errors. Spelling errors occur often in MultiNLI data, due to involvement of Turkers and noisy source text \cite{ghaeini2018dr}, which is problematic as some NLI systems rely heavily on word embeddings. Inspired by \newcite{belinkov2017synthetic}, we construct a stress test for ``spelling errors'' by performing two types of perturbations on a word sampled randomly from the hypothesis: random swap of adjacent characters within the word (for example, \emph{``I saw Tipper with him at teh movie.''}), and random substitution of a single alphabetical character with the character next to it on the English keyboard. For example, \emph{``Agencies have been further restricted and given less choice in selecting contractimg methods''}.
% \noindent
% \textbf{AdjSWAP:} Swap adjacent characters in a single word sampled randomly from the hypothesis. Example, \emph{"I saw Tipper with him at teh movie."}\\
% \noindent
% \textbf{KBSWAP:} Substitute a single alphabetical character randomly sampled from the hypothesis with the character next to it on the English keyboard. Example, \emph{'Agencies have been further restricted and given less choice in selecting contractimg methods.'}
% TODO: PUSH THIS TO RESULTS/SUPPLEMENTARY We additionally perform perturbations on only function words (conjunctions, pronouns and articles), and on only content words (nouns and adjectives \gn{what about verbs and adverbs?}) in the hypothesis to study the effects.
% We follow common linguistic standards, by treating the set of all prepositions, conjunctions, pronouns and articles as function words, and the set of all nouns and adjectives as content words.

%Study the effect of perturbing a stop word to a content word
%\subsection{Numerical Reasoning}
%\label{sec:numreas}
%\an{Introduce only one common subsection which details all annotations}

\section{Experiments}
\label{section:experiments}
\subsection{Experimental Setup}
%\gn{One comment: a very natural question is ``what if you generated stress-test examples using the training set and used these to train a model?'' What is your answer to this? I suspect that such a model would easily pass many (but not all) of the stress tests. What does this say about the validity of your method? I think it's worth having a discussion of this somewhere.}

\begin{table*}
\centering
\resizebox{\textwidth}{!}{%
\begin{tabular}{|c|c c|c c|c|c c|c c|c c|c c|}
\hline & \multicolumn{2}{|c|}{\textbf{Original}} & \multicolumn{3}{|c|}{\textbf{Competence Test}} & \multicolumn{6}{|c|}{\textbf{Distraction Test}} & \multicolumn{2}{|c|}{\textbf{Noise Test}}\\ \cline{4-14}
& \multicolumn{2}{|c|}{\textbf{MultiNLI}} & \multicolumn{2}{|c|}{\textbf{}} & & \multicolumn{2}{|c|}{\textbf{Word}} & \multicolumn{2}{|c|}{\textbf{}}  & \multicolumn{2}{|c|}{\textbf{Length}} & \multicolumn{2}{|c|}{\textbf{Spelling}} \\
\bf System & \multicolumn{2}{|c|}{\textbf{Dev}} & \multicolumn{2}{|c|}{\textbf{Antonymy}} & \textbf{Numerical} & \multicolumn{2}{|c|}{\textbf{Overlap}} & \multicolumn{2}{|c|}{\textbf{Negation}}  & \multicolumn{2}{|c|}{\textbf{Mismatch}} & \multicolumn{2}{|c|}{\textbf{Error}} \\ \cline{2-5} \cline{7-14}
\bf & \bf Mat & \bf Mis & \bf Mat & \bf Mis & \textbf{Reasoning} & \bf Mat & \bf Mis & \bf Mat & \bf Mis & \bf Mat & \bf Mis & \bf Mat & \bf Mis \\ \hline
\textbf{NB} & 74.2 & 74.8 & 15.1 & 19.3 & 21.2 & 47.2 & 47.1 & 39.5 & 40.0 & 48.2 & 47.3 & 51.1 & 49.8 \\ \hline
\textbf{CH} & 73.7 & 72.8 & 11.6 & 9.3 & 30.3 & 58.3 & 58.4 & 52.4 & 52.2 & 63.7 & 65.0 & 68.3 & 69.1 \\ \hline
\textbf{RC} & 71.3 & 71.6 & 36.4 & 32.8 & 30.2  & 53.7 & 54.4 & 49.5 & 50.4 & 48.6 & 49.6 & 66.6 & 67.0 \\ \hline
\textbf{IS} & 70.3 & 70.6 & 14.4 & 10.2 & 28.8 & 50.0 & 50.2 & 46.8 & 46.6 & 58.7 & 59.4 & 58.3 & 59.4 \\ \hline
\textbf{BiLSTM} & 70.2 & 70.8 & 13.2 & 9.8 & 31.3 & 57.0 & 58.5 & 51.4 & 51.9 & 49.7 & 51.2 & 65.0 & 65.1 \\ \hline
\textbf{CBOW} & 63.5 & 64.2 & 6.3 & 3.6 & 30.3 & 53.6 & 55.6 & 43.7 & 44.2 & 48.0 & 49.3 & 60.3 & 60.6\\ \hline
\end{tabular}
}
\caption{Classification accuracy (\%) of state-of-the-art models on our constructed stress tests. Accuracies shown on both genre-matched and mismatched categories for each stress set. For reference, random baseline accuracy is 33\%.}
\label{tab:accuracy}
\end{table*}

We focus on the following sentence-encoder models, which achieve strong performance on MultiNLI:

\noindent
\textbf{\newcite{nie-bansal:2017:RepEval} (NB)}: This model uses a sentence encoder consisting of stacked BiLSTM-RNNs with shortcut connections and fine-tuning of embeddings. It achieves the top non-ensemble result in the RepEval-2017 shared task \cite{nangia-EtAl:2017:RepEval}.\\%and an MLP for classification
\textbf{\newcite{chen-EtAl:2017:RepEval} (CH)}: This model also uses a sentence encoder consisting of stacked BiLSTM-RNNs with shortcut connections. Additionally, it makes use of character-composition word embeddings learned via CNNs, intra-sentence gated attention and ensembling to achieve the best overall result in the RepEval-2017 shared task.\\
\textbf{ \newcite{balazs-EtAl:2017:RepEval} (RiverCorners - RC)}: This model uses a single-layer BiLSTM with mean pooling and intra-sentence attention. \\
\textbf{\newcite{conneau-EtAl:2017:EMNLP2017} (InferSent - IS)}: This model uses a single-layer BiLSTM-RNN with max-pooling. It is shown to learn robust universal sentence representations which transfer well across several inference tasks.\\% with an MLP for classification
We also set up two simple baseline models:\\
\noindent
\textbf{BiLSTM:} The simple BiLSTM baseline model described by \newcite{nangia-EtAl:2017:RepEval}. \\%   \\
\textbf{CBOW:} A bag-of-words sentence representation from word embeddings. \\

\subsection{Model Performance on Stress Tests}
% Our stress tests seek to benchmark the performance of these models on six challenging linguistic phenomena, to highlight which phenomena remain challenging for current models and which phenomena can be handled. Table \ref{tab:accuracy} shows the performance (classification accuracy) of all six models on our stress tests as well as the original MultiNLI development set.
%, suggesting that simpler architectures might also be robust to such tests

Table \ref{tab:accuracy} shows the classification accuracy of all six models on our stress tests and the original MultiNLI development set. We see that performance of all models drops across all stress tests. On competence stress tests, no model is a clear winner, with \textbf{RC} and \textbf{CH} performing best on antonymy and numerical reasoning respectively. On distraction tests, \textbf{CH} is the best-performing model, suggesting that their gated-attention mechanism handles shallow word-level distractions to some extent. Interestingly, our \textbf{BiLSTM} baseline is the second-best model on two out of three distraction tests. On the noise test, \textbf{CH}, \textbf{RC} and both baselines \textbf{[BiLSTM;CBOW]} do not show much performance degradation, most likely due to the benefit of subword modeling via character-CNNs and the use of mean pooling. We further analyze model performance on each class of tests. 

% We perform further analysis of model performance to investigate which techniques help models handle certain phenomena. % and what kinds of errors these models make on each stress test. 

\subsection{Model Competence}
\noindent
\textbf{Model Performance on Antonymy:}
%\an{\textbf{1734 matched and 1561 mismatched}}
%\subsubsection{Antonyms}
Table \ref{tab:accuracy} shows that all models perform poorly on antonymy. \textbf{RC} achieves the best performance, with 36.4\% and 32.8\% on matched and mismatched sets respectively which is just higher than random performance. 
%Table \ref{tab:anterror} shows the percentage of C-E (contradiction predicted as entailment) and C-N (contradiction predicted as neutral) errors that all the models make on the antonymy stress test.\footnote{Note that since all examples in the antonymy test have the gold label ``contradiction'', models can make only C-E and C-N errors}
Our analysis shows that models tend to overpredict entailment (due to a high amount of word overlap in this test). This accounts for, on average, 86.4\% and 87.6\% of total errors on matched and mismatched sets.\footnote{Detailed results from this analysis are provided in appendix B} 

We study which antonym pairs are easy and difficult for models by examining the errors of the best and worst performing models on this test \textbf{[RC;CH]}. On 982 samples where both models fail, we find 617 unique antonym pairs, and on 171 samples where both models succeed, we find 84 unique antonym pairs. 89.8\% of the ``easy'' and 57.2\% of the ``hard'' antonym pairs appear in a contradiction relation within the training data, suggesting that models succeed on easy antonym-pairs seen in the training data but struggle to generalize. 

We were also curious about error variation by antonym type. We randomly sample 100 examples where both models fail and 100 samples where both succeed, and manually annotate whether the antonym present was gradable, relational or complementary. Among successful examples, 99\% are complementary antonyms with only one relational antonym. Amongst the failure cases, 20\% are relational antonym pairs, 73\% are complementary and 7\% are gradable, suggesting that models find relational and gradable antonyms hard, but get complementary antonyms both right and wrong. Finally, we examine differences between models by analyzing examples classified correctly by the best model which are not handled by the worst. We find that antonym pairs recognized by the weaker model occur, on average, nearly twice as often in the training data as antonym pairs recognized by the stronger model, suggesting that \textbf{RC} is able to learn antonymy from fewer examples (though these examples must be present in training data).\\
% This indicates that infusing models with linguistic insight might help in handling of phenomena such as antonymy. \an{can we find a way to empirically/ qualitatively justify this?} 
%\begin{table}
%\centering
%\resizebox{\columnwidth}{!}{%
%\begin{tabular}{|c|c c|c c|}
%\hline \textbf{System} & \multicolumn{2}{|c|}{\textbf{C-E Errors}} & \multicolumn{2}{|c|}{\textbf{C-N Errors}} \\ \cline{2-5}
%& \bf Mat & \bf Mis & \bf Mat & \bf Mis \\ \hline
%\textbf{NB} & 79.83 & 82.40 & 20.17 & 17.60 \\ \hline
%\textbf{CH}& 99.78 & 99.75 & 0.22 & 0.25 \\ \hline
%\textbf{RC} & 66.67 & 68.50 & 33.33 & 31.50 \\ \hline
%\textbf{IS} & 99.40 & 99.81 & 0.60 & 0.19 \\ \hline
%\textbf{B1} & & & & \\ \hline
%\textbf{B2} & & & & \\ \hline
%\end{tabular}
%}
%\caption{Percentage of C-E and C-N errors on antonymy test}
%\label{tab:anterror}
%\end{table}
% \begin{itemize}
% \item What kinds of errors do models make?
% \item Errors by length
% \item Errors by position of antonym in sentence
% \item Errors by type of antonym
% \item Errors by centrality of antonym
% \item what antonym pairs are always easy.., and what are always difficult?
% Moreover, none of the systems compute asymmetric representations for premise-hypothesis combination, further compounding these errors. 
% \end{itemize}

\noindent
\textbf{Model Performance on Numerical Reasoning:} Table \ref{tab:accuracy} shows that all models exhibit a significant performance drop on numerical reasoning, with none achieving an accuracy better than random (33\%). We analyze the predictions of the best and worst performing models on this test \textbf{[BiLSTM;NB]}. The biggest source of common errors for both models (1703 out of 4337 errors) is misclassifying neutral pairs as entailment, which arises because our construction technique flips entailed premise-hypothesis pairs to create neutral pairs, leading to high word overlap for neutral pairs. Our constructions also lead to high word overlap for contradiction pairs, leading to a large number of C-E errors for both models (1695 out of 4337 errors). Thus, 78.3\% of all errors are caused due to the models falsely predicting entailment.  Most of the remaining errors are caused by entailment examples containing the phrases ``more than'' or ``less than'' being incorrectly classified as contradiction. This behavior could arise as these phrases are often used by crowdworkers to create contradictory examples in the original MultiNLI data, fooling models into marking examples with this phrase as ``contradiction'' without reasoning about involved quantities. Our observations suggest that models do not perform quantitative reasoning, but rely on word overlap and other shallow lexical cues for prediction.  
%An interesting point to note is that due to our construction techniques, premise-hypothesis pairs for entailment and contradiction have high word overlap, leading to a large number of C-E (contradiction predicted as entailment) errors for all models (79.71\% of the errors committed by \textbf{CH}, the best performing model on this test are C-E errors). We analyze the predictions of the best and worst performing models on this test \textbf{[CH;NB]}. We find that both models rely highly on lexical cues, rather than quantitative reasoning, to predict contradiction. Out of 1266 contradiction examples in our test in which premise and hypothesis sentences are exactly same, with the exception of one number which has been changed, \textbf{CH} is able to classify 35 correctly, while \textbf{NB} is able to classify 75 correctly, indicating that both models ignore numerical differences. On the other hand, most entailment examples incorrectly classified as contradiction contain the phrase ``less than''. This behavior could arise as this phrase is often used by crowdworkers to create contradictory examples in the original MultiNLI data, fooling models into marking examples with this phrase as ``contradiction'' without reasoning about involved quantities. Our observations suggest that models do not perform quantitative reasoning, but rely on shallow lexical cues for prediction.

\subsection{Model Distraction}
Our distraction tests are designed to check model robustness to: 1) decreasing lexical similarity between premise-hypothesis pairs, 2) strong negation words in sentence pairs.\\

\noindent
\textbf{Effect of Decreasing Lexical Similarity:} Due to our construction methodology (appending tautologies), accuracy on \emph{word overlap} and \emph{length mismatch} demonstrates the effect of decreasing lexical similarity on model performance. Table \ref{tab:accuracy}, shows accuracy decreases for all models on both tests. This drop is lower for \textbf{CH}, suggesting that their gated attention mechanism might help in focusing on relevant parts of the sentence.

The significant decrease in accuracy indicates that lexical similarity is a strong signal for entailment prediction, failing which models default to predicting neutral. To provide further justification, we compare the proportion of false neutral errors for all models on word overlap and length mismatch stress sets vs. the original MultiNLI development set. As shown in Table \ref{tab:woerror}, we find it increases for all models on both sets.\\
\begin{table}
\parbox{.45\linewidth}{
\centering
\begin{tabular}{|c|c c|c c|c c|}
\hline \textbf{Sys} & \multicolumn{2}{|c|}{\textbf{MultiNLI}} & \multicolumn{2}{|c|}{\textbf{Word}} & \multicolumn{2}{|c|}{\textbf{Length}} \\
\textbf{tem}& \multicolumn{2}{|c|}{\textbf{Dev}} & \multicolumn{2}{|c|}{\textbf{Overlap}} & \multicolumn{2}{|c|}{\textbf{Mismatch}} \\ \cline{2-7}
& \bf Mat & \bf Mis & \bf Mat & \bf Mis & \bf Mat & \bf Mis \\ \hline
\textbf{NB} & 33.2 & 33.1 & 43.2 & 38.3 & 46.0 & 46.9 \\ \hline
\textbf{CH} & 32.9 & 31.7 & 84.7 & 85.3 & 65.8 & 65.8 \\ \hline
\textbf{RC}& 37.1 & 39.1 & 74.3 & 83.3 & 74.2 & 79.5 \\ \hline
\textbf{IS} & 34.7 & 31.4 &  86.3 &  87.0 & 43.5 & 44.2 \\ \hline
\textbf{BiLSTM} & 38.5 & 37.9 & 83.2 & 81.9 & 75.9 & 79.1 \\ \hline
\textbf{CBOW} & 33.9 & 30.2 & 74.5 & 72.3 & 54.7 & 59.9 \\ \hline
\end{tabular}
\caption{ \% of \textsc{false neutral} in total errors on MultiNLI development set, word overlap test and length mismatch test.}
\label{tab:woerror}
}
\hfill\parbox{.45\linewidth}{
\begin{tabular}{|c|c c|c c|}
\hline \textbf{System} & \multicolumn{2}{|c|}{\textbf{BiLSTM}} & \multicolumn{2}{|c|}{\textbf{NB}} \\ \cline{2-5} 
& \bf Mat & \bf Mis & \bf Mat  & \bf Mis \\ \hline
\textbf{MultiNLI Dev} & 70.2 & 70.4 & 66.6 & 66.6  \\ 
\textbf{NEGATION} & 68.9 & 70.4  & 49.3 & 48.7 \\ 
\textbf{DIFF TAUT} & 49.0 & 49.3 & 49.9 & 49.7 \\ \hline
\end{tabular}
\caption{Effect of training on distraction data on original DEV set, original distraction set and new distraction set}
\label{tab:negerror}
}
\end{table}

\noindent
\textbf{Effect of Introducing Strong Negation Words:} 
% To study the effect of introducing strong negation words on model performance, we observe model accuracies on our negation stress test. 
Table \ref{tab:accuracy} shows results on negation, and we see that all state-of-the-art models perform poorly, with accuracies decreasing by 23.4\% and 23.38\%, on average, on matched and mismatched sets respectively. However, comparing the number of E-C (entailment predicted as contradiction) and N-C (neutral predicted as contradiction) errors for these models on the negation test vs. the original MultiNLI development set, we do not find an increase in these error types on negation. Instead, we observe an increase in false neutral errors for all models. This could occur due to the introduction of extra words (``false'', ``is'' and ``true'') apart from ``not'', indicating that decreasing lexical similarity has a stronger effect on models than introducing negation.

\subsection{Effect of Noise}
% We use our spelling error stress test to measure model robustness to random noisy perturbations in premise and hypothesis sentences.
Our noise test results in Table \ref{tab:accuracy} show that \textbf{NB} and \textbf{IS} exhibit a huge decrease in accuracy, since both models rely on word embeddings. Other models show little performance degradation on this test. \textbf{CH} performs subword modeling via character-level CNNs, which provides robustness towards perturbation attacks. \textbf{RC} and \textbf{BiLSTM} perform well despite relying on word embeddings since both use mean pooling, which might reduce the effect of single-word edits on the final representation. \textbf{CBOW} is also very robust to this test, which can arise from the fact that it sums word embeddings to create the final sentence embedding, diluting the effect of changing a single word on final model performance\footnote{We analyze difference in model performance across perturbation techniques such as adjacent character swapping, keyboard character swapping, function word and content word perturbations, but do not observe significant differences. Results from these experiments are included in appendix C.}.
\subsection{Training with Distraction}
Finally, we study the effect of training with distractions generated via our adversarial construction. We generate an equivalent sample with the negation distraction for every sample in the training data, and retrain \textbf{NB} and \textbf{BiLSTM} on the union of these examples and original training data. 

%\textbf{FLIP TAUT} & 61.82 &  & 48.94 & 48.20 \\ \hline
%\textbf{BEG TAUT} & 67.86 & & 49.71 & 49.25 \\ \hline
We observe the performance of the trained models on three tests: the original MultiNLI development set, the negation stress test and a new distraction test creating using a different tautology \emph{``green is not red''} (\textsc{diff taut}). We observe that \textbf{NB} shows performance degradation across all tests, but training \textbf{BiLSTM} on distraction data helps it become robust to the tautology it was trained on. However, it collapses when evaluated on a different tautology. Ignoring such distractions is something humans do naturally. Models should not have to train on the specific distraction to succeed on this evaluation.
\section{Related Work and Discussion}
\label{ref:Discussion}

%\gn{This is pretty verbose. Could you try to condense things a bit more, grouping together similar papers so you don't need to have one sentence per paper?}

% \gn{Cite: \newcite{smith2012adversarial} and the papers it cites, other papers that cite it. Papers from this workshop and their references: \url{https://generalizablenlp.weebly.com/}.} 
%Aside from the differences noted in \ref{section:Introduction} It is also interesting to note that in our propositional framework we construct adversarial examples for natural language inference which share \emph{no lexical similarity} with the data, and yet are distracting to most NLI inference systems. %In this work, we propose such an evaluation scheme where models are ```stress tested``` on identified difficult phenomena in textual entailment.
Adversarial evaluation schemes have been proposed to evaluate model robustness on various NLP tasks. \newcite{smith2012adversarial} discuss dangers of community-wide ``overfitting'' to benchmark datasets and emphasize the need to correlate model errors to well-defined linguistic phenomena to understand specific model strengths and weaknesses. Prior work \cite{rimell2009unbounded,schneider-EtAl:2017:BLGNLP2017} performed analyses of model errors in dependency parsing and information extraction. Motivated by this desideratum, we analyze errors in Multi-Genre NLI and automatically construct large stress sets to evaluate NLI systems on identified difficulties. This is analogous to recent work \cite{jia-liang:2017:EMNLP2017,burlot2017evaluating} on developing automated adversarial evaluation schemes for reading comprehension and machine translation. However, unlike these efforts, our stress tests allow us to study model performance on a range of linguistic phenomena. Unlike work on manual construction of small adversarial evaluation sets for various NLP tasks \cite{levesque2014our,mahler-EtAl:2017:BLGNLP2017,staliunaite2017breaking,isabelle-cherry-foster:2017:EMNLP2017,belinkov2017synthetic,bawden2017evaluating}, our work focuses on a more exhaustive large-scale evaluation for NLI.  It is interesting to note that \newcite{bayer2006evaluating} discuss the daunting cost of finding entailment pairs for NLI evaluation, but our techniques can be used to construct such pairs with low cost.  
%While this provides some insight into model weaknesses, it does not tell us whether there are other similar examples the model works on (indicating if a model is weak against an entire category of errors, or just one example).  
%\newcite{} propose the Winograd Schema challenge, where systems attempt to perform coreference resolution on manually constructed examples. 
%\newcite{} manually construct small test sets to test four capabilities of machine translation systems. Similarly \cite{} evaluate machine translation systems on robustness to noise, discourse phenomena and morphological competence respectively. Such evaluation methodologies are a step in the right direction towards systematic evaluation. However, small manually constructed test sets might give an incomplete picture of a model's ability to handle specific linguistic phenomena. In our work we focus on large-scale phenomenon-by-phenomenon evaluation for natural language inference. %\newcite{} show that noise can have significant effect on model performance in Machine Translation. Inspired by this work, we examine the effect of synthetic noise in the form of grammaticality perturbations.

%Future ditand offer our corpus and software to support future work in the area.
%than on the basis of hypothesizing which phenomena would be challenging for NLI systems
The NLI task attracted significant interest before datasets became large enough for the application of neural methods \cite{glickman2005web,harabagiu2006methods,romanoinvestigating,dagan2006pascal,giampiccolo2007third,dagan2010fourth,maccartney2009natural,zanzotto2006learning,malakasiotis2007learning,haghighi2005robust,angeli2014naturalli}. \newcite{de2009multi} analyze the effect of multi-word expressions and find that they do not significantly affect the performance of NLI systems. Perhaps the closest to our contribution are the works of \newcite{cooper1996using}, which manually constructs sentences containing phenomena that NLI systems are expected to handle (FraCaS), and \newcite{marelli2014sick}, which constructs sentences that require compositional knowledge (SICK).  Our constructions differ from SICK and FraCaS in several aspects. Since we use large datasets \cite{williams2017broad,ling2017program} as base data, our sets are larger and more lexically diverse than both SICK (which used a seed set of 1500 sentences) and FraCaS (which was manually constructed). Secondly, while SICK uses handcrafted rules and incorporates linguistic phenomena, sentence-pairs are not constrained to exhibit only one phenomenon, which may introduce confounding factors during analysis. Though FraCaS follows the constraint of restricting sentence-pairs to exhibit only one phenomenon, it contains very few examples of each phenomenon. Conversely, our techniques generate large evaluation sets, with each set focusing on a single phenomenon, providing a testbed for fine-grained evaluation and analysis. Lastly, our stress tests are grounded in failings of current state-of-the-art models, rather than on phenomena hypothesized to be challenging for NLI models. Our evaluation sets also differ from the small portion of the MultiNLI development set annotated for challenging linguistic phenomena, as similar to SICK, each sentence pair is not constrained to exhibit a single phenomenon. In addition, presence of biases in the MultiNLI development and test data \cite{gururangan2018annotation,poliak2018hypothesis} could also lead to models exploiting them as shallow cues for prediction (for example, the performance of baseline models on the subset of the MultiNLI dataset annotated for antonymy averages 67\% , while the same baselines perform much worse on our antonymy stress-test). 

%Though FraCaS follows the constraint of restricting sentence-pairs to exhibit only one phenomenon, it contains very few examples of each phenomenon. Conversely, our techniques generate large evaluation sets, with each set focusing on a single phenomenon, providing a testbed for fine-grained evaluation and analysis.
%Our stress tests, however, specifically target identified weaknesses of neural models to safeguard against their pattern-matching behaviors, and offer fine-grained evaluation on the same phenomena.
%and the work of \newcite{cooper1996using} on the FraCaS dataset identifies phenomena that a NLI system should be able to handle
% The work of  identifies phenomena that a NLI system should be able to handle. However, though the dataset offers fine-grained evaluation, it requires examples to be manually constructed, making it difficult to construct large-scale suites of examples. In addition, our stress tests are meant to safeguard against identified pattern-matching behaviors of current neural models, which is becoming increasingly relevant with the advent of deep-learning techniques for natural language inference.

We hope that insights derived from our stress tests will stimulate future research in NLI. One promising direction would be the development and investigation of more linguistically-motivated neural models on NLI (such as models which incorporate explicit negation scope information or semantic roles for example), as our stress tests now provide a framework for in-depth analysis of model performance and demonstrate significant room for improvement in these areas. While we benchmark the performance of state-of-the-art models on our stress tests, in the future, it would be interesting to investigate which architectural choices contribute to model successes and \emph{why}. Another interesting research direction is the identification of ``core competencies'', such as quantitative reasoning or antonymy, which can enhance model performance across multiple NLP tasks such as sentiment analysis, question answering and relation extraction and studying transfer of representations from ``competent'' models.
\section{Conclusion}
%should emphasize construction
%In this work, we present a set of techniques, including a novel propositional logic framework, for automatic construction of a suite of large stress tests for targeted evaluation of NLI models on specific linguistic phenomena.
In this work, we present a suite of large-scale stress tests to perform targeted evaluation of NLI models, along with a set of techniques for their automatic construction. Our stress tests evaluate a model's ability to reason about quantities and antonymy (\emph{competence tests}), its susceptibility to shallow lexical cues (\emph{distraction tests}) and its robustness to random perturbations (\emph{noise tests}). We benchmark the performance of four state-of-the-art sentence encoding models on our tests and find that they struggle on many phenomena, despite reporting high accuracy on NLI. 
%Interestingly we observe that state-of-the-art models do not always outperform other models, with weaker models (BiLSTM baseline, \newcite{balazs}) performing better on some classes of tests.

% Our stress test evaluation also provides interesting insights about model strengths. For example, we observe that subword modeling via character-CNNs and mean-pooling provide robustness to spelling errors, while the gated-attention mechanism proposed by \newcite{chenRepEval} introduces some robustness to shallow lexical distractions. 
Overall, we consider the MultiNLI dataset to be a valuable resource for the NLP community, with entailment pairs drawn from several different genres of text. However, we argue that the community would benefit by having NLI models pass sanity checks, in the form of ``stress tests'', to ensure models evolve against exploiting simple idiosyncrasies of training data. Similar to \newcite{isabelle-cherry-foster:2017:EMNLP2017}, we intend our stress tests to supplement existing NLI evaluation rather than replace it. In the future, we hope benchmarking model performance on stress tests in addition to standard evaluation criteria will provide deeper insight into model strengths and weaknesses, and guide more informed model choices. We would also like to note that the ``stress test'' evaluation paradigm that we propose for NLI can be further updated in the future when new forms of models are devised, increasing the coverage of the tests to cover problems of future models as well. We release all our stress tests and associated resources to the community to promote work on models that get us closer to true natural language understanding. 
\section*{Acknowledgements}
This work has partially been supported by the National Science
Foundation under Grant
No. CNS 13-30596. The views and conclusions
contained herein are those of the authors and
should not be interpreted as necessarily representing
the official policies or endorsements, either expressed
or implied, of the NSF, or the US Government. This work has also been supported through the CMLH Fellowship in Digital Health for the author Naik. The authors would also like to thank Shruti Rijhwani, Siddharth Dalmia, Shruti Palaskar, Khyathi Chandu, Aditya Chandrasekar, Paul Michel, Varshini Ramaseshan and Rajat Kulshreshtha for helpful discussion and feedback with various aspects of this work.
\bibliographystyle{acl}
\bibliography{coling2018}

\section*{Appendix A. Error Analysis on Mismatched Set}
Table \ref{tab:mismaterror} describes what proportion of misclassified examples from the mismatched set fall into each error category, as defined in the typology presented in \S\ref{section:NLI}.
\begin{table}[h]
\centering
\begin{tabular}{|c|c|}
\hline \textbf{Error Category} & \textbf{\% of Misclassified Examples } \\ \hline
\textbf{Word Overlap} & 28\\
\textbf{Negation} & 12\\
\textbf{Antonymy} & 4\\
\textbf{Numerical Reasoning} & 4\\
\textbf{Length Mismatch} & 6\\
\textbf{Grammaticality} & 4\\
\textbf{Real-World Knowledge} & 14\\
\textbf{Ambiguity} & 8\\
\textbf{Unknown} & 20\\ \hline
\end{tabular}
\caption{Distribution of misclassified examples from the MultiNLI mismatched development set.}
\label{tab:mismaterror}
\end{table}

\section*{Appendix B. Error Types on Antonymy}
The high amount of word overlap in this test causes models to overpredict entailment, accounting for, on average, 86.4\% and 87.6\% of total errors on matched and mismatched sets respectively. We present the exact proportion of  False Entailment and False Neutral errors in Table \ref{tab:anterror}. Keep in mind there is only one gold class in this category: contradiction.

\begin{table}[h]
\centering
\begin{tabular}{|c|c c|c c|}
\hline \textbf{System} & \multicolumn{2}{|c|}{\textbf{C-E Errors}} & \multicolumn{2}{|c|}{\textbf{C-N Errors}} \\ \cline{2-5}
& \bf Mat & \bf Mis & \bf Mat & \bf Mis \\ \hline
\textbf{NB} & 79.83 & 82.40 & 20.17 & 17.60 \\ \hline
\textbf{CH}& 99.78 & 99.75 & 0.22 & 0.25 \\ \hline
\textbf{RC} & 66.67 & 68.50 & 33.33 & 31.50 \\ \hline
\textbf{IS} & 99.40 & 99.81 & 0.60 & 0.19 \\ \hline
\end{tabular}
\caption{Percentage of C-E and C-N errors on antonymy test.}
\label{tab:anterror}
\end{table}

As expected, all four models make a high amount of false entailment errors because they notice high amounts of lexical similarity between the premise and the hypothesis. 
\section*{Appendix C. Additional Experiments with Spelling Error Stress Tests}
This stress test consists of an adversarial example set which tests model robustness to spelling errors. Spelling errors occur often in MultiNLI data, due to involvement of Turkers and noisy source text, which is problematic as some NLI systems rely heavily on word embeddings. We construct a stress test for ``spelling errors'' by performing two types of perturbations:\\
\noindent
\textbf{AdjSWAP:} Swap adjacent characters in a single word sampled randomly from the hypothesis. For example, \emph{``I saw Tipper with him at teh movie."}.\\
\noindent
\textbf{KBSWAP:} Substitute a single alphabetical character randomly sampled from the hypothesis with the character next to it on the English keyboard. For example, \emph{``Agencies have been further restricted and given less choice in selecting contractimg methods.''}
We additionally perform perturbations on only function words (conjunctions, pronouns and articles), and on only content words (nouns and adjectives) in the hypothesis to study the effects. We do not address perturbations in verbs and adverbs in the content word vs. function word analysis. The results are presented in Table \ref{tab:gram}.

\begin{table} [h]
\small
\centering
\begin{tabular}{|c|c c |c c |c c |c c |}
\hline \bf Sys& \multicolumn{2}{|c|}{\bf \textsc{Adj SWAP}} & \multicolumn{2}{|c|}{\bf \textsc{KB Swap}} & \multicolumn{2}{|c|}{\bf \textsc{CN Swap}} & \multicolumn{2}{|c|}{\bf \textsc{FN Swap}} \\
\bf tem & \bf Mat & \bf Mis & \bf Mat & \bf Mis & \bf Mat & \bf Mis & \bf Mat & \bf Mis \\ \hline
\textbf{NB} & 43.0 & 42.9 & 47.7  & 47.9 & 51.1 & 49.8 & 49.7 & 49.6 \\ \hline
\textbf{CH} & 68.24 & 68.1 & 68.5 & 68.3 & 68.3 & 69.1 & 69.9 & 70.3 \\ \hline
\textbf{RC} & 66.6 & 66.4 & 67.0 & 66.8 & 66.6 & 67.0 & 68.4 & 68.4 \\ \hline
\textbf{IS} & 57.8 & 58.6 & 57.7 & 58.7 & 58.3 & 59.4 & 57.5 & 57.6 \\ \hline
\end{tabular}
\caption{Model Performance on grammaticality test}
\label{tab:gram} 
\end{table}

We observe that there is no significant effect of perturbing a function word or a content word. One hypothesis is that content words can often be named entities for which the models do not find word embeddings. We also do not find a considerable difference in performance between the different kinds of perturbations but this is expected behaviour as most models use word embeddings, and these will just be categorized as unknown words.

\end{document}